\documentclass[letterpaper, 10 pt, conference]{ieeeconf}  % Comment this line out if you need a4paper

\usepackage{textcomp}
\usepackage{color,soul}
\usepackage{graphicx}
\usepackage{multirow}
\usepackage{hhline}
\let\labelindent\relax
\usepackage[inline]{enumitem}
\usepackage{float}
\usepackage{amsmath}
\usepackage[ruled,vlined]{algorithm2e}
\usepackage[table,x11names]{xcolor}
\usepackage{tabularx}
\usepackage{booktabs}
\usepackage{flushend}
\usepackage{siunitx}
\usepackage{hyperref}
\usepackage[capitalise]{cleveref}

\newcolumntype{Y}{>{\centering\arraybackslash}X}

\newcommand{\ie}{i.e. }
\newcommand{\eg}{e.g. }

\SetKwInput{KwInput}{Input}                % Set the Input
\SetKwInput{KwOutput}{Output}              % set the Output

\IEEEoverridecommandlockouts                              % This command is only needed if 
\crefname{section}{Sec.}{Sec.}

\title{\LARGE \bf
PointCrack3D: Crack Detection in Unstructured Environments using a 3D-Point-Cloud-Based Deep Neural Network$^*$
}

\author{Faris Azhari$^{1,2}$, Charlotte Sennersten$^{3}$, Michael Milford$^{1}$ and Thierry Peynot$^{1,2}$% <-this % stops a space
\thanks{*This work was supported by Australian Government Research Training Program (RTP), Queensland University of Technology (QUT) through the QUT Centre for Robotics and Mining3.}% <-this % stops a space
\thanks{$^{1}$QUT Centre for Robotics, Brisbane, Australia.
        {\tt\small faris.azhari@connect.qut.edu.au} }%
\thanks{$^{2}$Mining3, Pinjarra Hills, Australia.}%
\thanks{$^{3}$CSIRO Mineral Resources, Pullenvale, Australia.}
}

\begin{document}

\maketitle
\thispagestyle{empty}
\pagestyle{empty}

%%%%%%%%%%%%%%%%%%%%%%%%%%%%%%%%%%%%%%%%%%%%%%%%%%%%%%%%%%%%%%%%%%%%%%%%%%%%%%%%

\begin{abstract}

Surface cracks on buildings, natural walls and underground mine tunnels can indicate serious structural integrity issues that threaten the safety of the structure and people in the environment. Timely detection and monitoring of cracks is crucial to managing these risks, especially if the systems can be made highly automated through robots. Vision-based crack detection algorithms using deep neural networks have exhibited promise for structured surfaces such as walls or civil engineering tunnels, but little work has addressed highly unstructured environments such as rock cliffs and bare mining tunnels. To address this challenge, this paper presents PointCrack3D, a new  3D-point-cloud-based crack detection algorithm for unstructured surfaces. %using a deep neural network. 
The method comprises three key components: an adaptive down-sampling method that maintains sufficient crack point density, a DNN that classifies each point as crack or non-crack, and a post-processing clustering method that groups crack points into crack \emph{instances}. 
The method was validated experimentally on a new large natural rock dataset, comprising coloured LIDAR point clouds spanning more than 900~m\textsuperscript{2} and 412 individual cracks. 
Results demonstrate a crack detection rate of 97\% overall and 100\% for cracks with a maximum width of more than 3~cm, significantly outperforming the state of the art. 
Furthermore, for cross-validation, PointCrack3D was applied to an entirely new dataset acquired in different location and not used at all in training and shown to detect 100\% of its crack instances.
%Additionally, the trained PointCrack3D model was cross-validated on a second dataset collected from a different environment. 
%To aid further development, 
We also characterise the relationship between detection performance, crack width and number of points per crack, providing a foundation upon which to make decisions about both practical deployments and future research directions.

% The method is shown to successfully detect 97\% of the cracks in the test set and strongly outperforms a state-of-the-art point-cloud-based fault detection method. 
% %% TODO: add parameter configuration + LIDAR only.
% In addition, we provide an analysis of the detection capability in relation to crack width and number of 3D points, which provide insight into the future development of autonomous crack detection robots in unstructured environments. 

% A comparison to a state-of-the-art point-cloud-based fault detection is made and our proposed method performed 30\% better in absolute in terms of detecting instances of cracks.  

\end{abstract}
\section{Introduction}

Cracks on structures, whether human-made (\eg pavements, buildings, bridges, tunnels) or natural (\eg cliffs, rock surfaces, caves), can be defined as the structure that breaks a surface continuity with a gap.
They are signs of potential weakness, and can ultimately lead to catastrophic disasters such as collapses and rockfalls. For example, in an underground mining environment, surface cracks (also known as fractures) are one of the early signs of geological failures that can lead to rockfalls or even collapse of the roof~\cite{molinda2000,szwedzicki2003}. Over a 10-year period (2010 - 2019), in the U.S.A. alone, 49 deaths and 3,359 non-fatal accidents were reported as a result of geological failures~\cite{msha2020}. This toll accounts for over 25\% of the total death toll in underground mines. Current mitigation techniques include routine visual inspections that 
%by a geotechnical engineer to spot cracks and implement design-specific support systems. This inspection process 
pose a risk to the safety of the geotechnical engineer, especially when surveying voids that are abandoned, newly blasted or post-ore extraction. 
%Further risks include rough ground, darkness, airborne dust, harmful gas buildups, and extreme heat. 
Automated inspection systems using robots could eliminate this risk and increase survey frequency, resulting in earlier detection. This work focuses on the development of a key component of this system: automated crack detection. %including online real-time detection to guide further targeted crack mapping and characterisation.
This detector should be capable of: detecting crack \emph{instances} (so they can be counted), determining their position (globally and relatively to each other) and characterising them (e.g. estimating their width and length).

%For a crack detection application, it is important to be able to detect the existence of crack instances, whereby the detection of points making up cracks allows for the extraction of accurate measurements \ie width and length.

% Hence it is important not to only detect cracks on such structures but also to obtain the characteristics (i.e. position, width, length, etc.) of each individual cracks. Knowing the existence of cracks and their attributes allows for informed decisions to be made by engineers on the structure itself.
\begin{figure}[t]
    \centering
    \includegraphics[width=0.9\linewidth]{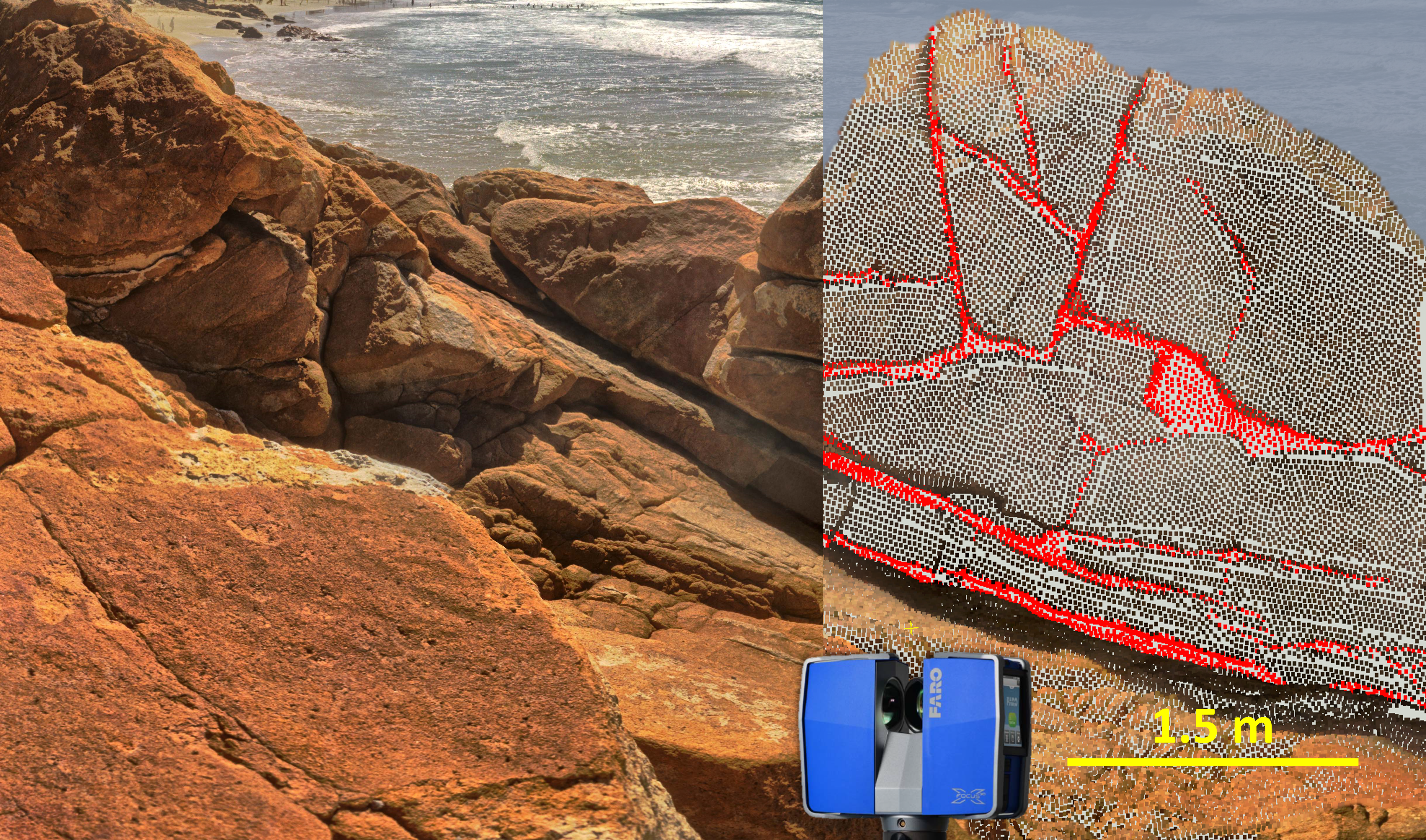}
    \caption{Scanning an unstructured surface with cracks using a 3D LIDAR scanner with the coloured point cloud and labelled crack (red) overlaid.}
    \label{fig:scanning}
\end{figure}

A large body of research has investigated the detection of cracks in structured environments, on mostly flat or smooth surface, using vision e.g.~\cite{Chen2020}. %% Add a couple of references from Sec. II
However, as noted in~\cite{azhari2019}, in imagery of an unstructured surface, shadows of a protruding structure can easily be confused with an actual crack. 
Prior work has shown that the performance of state-of-the-art (SOTA) image-based crack detection methods is significantly compromised when applied to a shotcrete-layered unstructured surface in an underground mine site~\cite{azhari2019}, even though the surfaces in this case remain relatively smooth thanks to the shotcrete. In this paper we aim to detect cracks in significantly more challenging scenes, such as highly unstructured bare rock surfaces, see~\cref{fig:scanning} and \cref{fig:outputpc}.

%An experimental study of the performance of state-of-the-art (SOTA) image-based crack detection methods on images of a shotcrete-layered unstructured surface in an underground mine site \cite{azhari2019} suggest that these methods performed significantly lower than on smooth human-made surfaces. However, shotcrete smooths out the unstructured surface underneath while this paper is concerned of detecting cracks on bare unstructured surfaces.

To address the challenge, the automated crack detection method proposed in this paper exploits LIDAR data to capture accurate geometric information. The properties of most cracks - a thin opening that covers a relatively small area on the surface - makes for a challenging problem as point clouds in the crack region can be sparser compared to neighbouring flat contiguous surfaces.
Inspired by promising early results obtained using DNN methods, this paper proposes PointCrack3D (PtCrack3D), a point-cloud-based DNN approach for crack detection on highly unstructured surfaces. 
%The DNN classifies each point as crack or non-crack.
%For a crack detection application, it is important to be able to detect the existence of crack instances, whereby the detection of points making up cracks allows for the extraction of accurate measurements \ie width and length.
PtCrack3D comprises three key components: an adaptive down-sampling method that maintains sufficient crack point density, a DNN that classifies each point as crack or non-crack, and a post-processing clustering method that groups crack points into crack \emph{instances}. 

The proposed method was trained and validated experimentally on a new dataset of a large natural rock cliff, on which it is shown to detect 97\% of the crack instances overall, and 100\% of the cracks at least 3~cm wide, thereby strongly outperforming a state-of-the-art point-cloud based defects detection method. 
Furthermore, the paper proposes a cross-validation of the method, where PtCrack3D applied `out of the box' is shown to detect 100\% of the crack instances of an entirely new dataset acquired in different location and not used at all in the training. 
%Cross-validating PtCrack3D on another natural rock dataset shows 100\% crack detection rate, further supporting PtCrack3D detection capabilities on an out-of-sample case.

The rest of the paper is structured as follows. 
\cref{sec:litreview} discusses the relevant literature. 
%on vision-based and point-cloud-based crack detection system, as well as DNN methods for point clouds.
\cref{sec:methods} outlines the crack detection method and implementation followed by
\cref{sec:results}, which evaluates the performance of the approach.
Finally, conclusions and future work are discussed in \cref{sec:conclusion}.

\section{Related Work}\label{sec:litreview}

The automation of crack detection has been a topic of wide interest that follows along with the advancement and affordability of perception sensors such as cameras and LIDARs. In addition, with the advancement of Deep Neural Network (DNN) methods, utilising DNN architectures has become the SOTA in detecting cracks~\cite{Chen2020,Dung2019,zhu2019,zou2019}. However, existing research has focused on detecting cracks on structured surface i.e. mostly flat or with a known geometry such as on pavements~\cite{Chen2020,zhu2019,zou2019}, concrete~\cite{Dung2019,zhu2019,Chen2020} or tunnels~\cite{Montero2017} using monocular vision. These vision-based methods have shown promising results, but have been seen to struggle when applied to more challenging environments where the surface is unstructured~\cite{azhari2019}. 3D sensors have also been utilised for crack and defects detection, however, they have been applied on similar structural domains such as pavements~\cite{Zhong2020,Zhang2019-crack}, concrete~\cite{Xu2019,sarker2017,Cho2018}, timber~\cite{cabaleiro2017} and aeroplane exterior~\cite{jovancevic2017}. 

To the best of the authors' knowledge no prior work achieved LIDAR-based crack detection method in an unstructured environment using deep learning. Hence, in this section, relevant literature discussed concerns:
\begin{enumerate*}[label={(\alph*)}]
    \item point cloud-based defects detection systems;
    \item DNN architecture for point cloud.
\end{enumerate*}

\subsection{Point cloud-based defects detection methods}

Point cloud-based crack detection has not been explored as much as its image-based counterpart. Hence, this section explores defects detection, a more general application domain. In \cite{Xu2019,Cho2018}, the point clouds of concrete surfaces are projected to a 2D plane producing range images and are used for detection, for example, applying DNN method on the images~\cite{Allen2018}. Other methods include grid search~\cite{Zhong2020} and plane fitting~\cite{cabaleiro2017} which have been applied on pavement and timber point clouds, respectively. Defects on aeroplane exteriors have been detected using region growing on the surface normal estimation~\cite{jovancevic2017}. In~\cite{madrigal2017}, Model Point Feature Histogram (MPFH) uses local surface descriptors paired with a machine learning method to detect defects on a variety of point cloud surfaces (\eg welding, artificial teeth, ceramics) where one of the defects are surface cracks. MPFH was reported to achieve an average accuracy of 94\% in the best configuration. %and is considered the state of art??? 
In these applications, the surfaces are flat or at least a priori known and could be easily modelled, unlike the highly unstructured surfaces in this work. 
%which may not work well when applied to unstructured surfaces.

\subsection{Deep neural network architecture for point cloud}

PointNet~\cite{Charles2017} was the first DNN method that performs convolution directly on the raw point cloud input without the need of any point cloud representation such as voxelisation or projection. Further works based on PointNet was done such as adding local region computation~\cite{charles2017-pp, Hua2018}. Based on local region computations,~\cite{Wu2019,li2018-pointcnn} added a module that computes the correlation between the extracted local regions and achieved better performance compared to methods without it. Additionally, implementations such as~\cite{randlanet,cylinandasym,fgnet} were proposed for fast large-scale point cloud segmentation. However, these methods were never tested on large imbalanced datasets in a unstructured environment where the object of interest is small compared to the scene and represented by points in low density.

Point cloud-based DNN methods have been successfully applied for semantic segmentation on real-world data, for example, segmenting ground, buildings and vegetation of aerial point cloud~\cite{Soilan2019}, indoor and outdoor object segmentation in~\cite{Li2020,Ma2019}, respectively. However, point cloud DNN has not been applied to segmentation problems with high class imbalance.
%where the representation of the object of interest is dominated by a large margin compared to the background points. 

\section{PointCrack3D}\label{sec:methods}

This paper proposes PointCrack3D, a point-cloud-based crack detection algorithm for unstructured surfaces using a DNN architecture. Given a point cloud of an unstructured surface, the objective is first to classify each point into either crack or non-crack and then to cluster crack points into their respective crack instances to perform crack instance detection. The system consists of three components:
\begin{enumerate*}[label={(\alph*)}]
    \item Data preparation;
    \item DNN architecture;
    \item Point clustering.
\end{enumerate*} 

\subsection{Pre-Processing}
Given a point cloud of a scanned surface, the raw data is first discretised into voxels. The point clouds contained in each voxel are then sampled down to meet the required DNN input resolution followed by normalisation of the point coordinates and its corresponding feature set.

The data preparation process is governed by the parameter $[d,n,s]$, where $d$ is the dimension of the voxel, $n$ is the number of points inside a voxel and $s$ is the stride at which the voxel moves in space, where $s=d$ means no overlapping between neighbouring voxels. Each voxel that exceeds $n$ number of points is then down-sampled using an adapted voxel grid-based method as shown in Algorithm~\ref{algo:downsample}. Respective to the nature of cracks points on a surface point cloud, this down-sampling method preserves the sparse points that represent a crack compared to applying random sampling method which leads to the diminishing of sparse crack points. This discretisation step (i.e. voxelisation followed by down-sampling) preserves the geometrical features of cracks on a large surface compared to using a whole down-sampled point cloud directly as an input. 

Finally the voxelised points and their respective features are normalised within the range of [0,1] with respect to the entire training dataset which finally gives the input as:
\begin{equation}
    \textstyle V_{i=1...v_n}=\textstyle P_j\{X_j,F_{j}\}_{j=1...n}
\end{equation} 
\noindent where $v_n$ is the number of voxels, $X_j$ and $F_j$ are the normalised coordinate, $\{x_j, y_j, z_j\}$ and the normalised features of the $j$-th points, and $n$ is the number of points in a voxel. This gives an input size of $n \times m$ where $m$ depends on the dimension of features included, $m=3+dim(F)$.
\begin{algorithm}[ht]
    \SetAlgoLined
    \KwInput{Point cloud = \(pc\), No. desired points = \(n\)}
    \KwOutput{Down-sampled \(pc = pc_{out}\)}
    \(grid = floor(\sqrt[3]{n})\)\;
    \(v = voxelise(pc, grid)\)\;
    \While{\(size(v) < n\)}{
        \(grid\mathrel{{+}{+}}\)\;
        \(v = voxelise(pc, grid)\)\;
    }
    \If{\(size(v) > n\)}{
        \(v_{s} =\) choose \([size(v)-n]\)  randomly from \(v\)\;
        \ForEach{\(i \in index(v_{s})\)}{
            \(v_{n}\) = choose a random voxel neighbour of \([v_{s}(i) | \notin v_{s}]\)\;
            \(v_{s}(i) = [v_{s}(i) \ ; \ v_{n}]\)\;
        }
    }
    \ForEach{\(i \in index(v)\)}{
        \(c = mean(v_{pc}(i))\)\;
        \(pc_{out}(i) = min(dist(v_{pc}(i), c))\)\;
    }
    \caption{Modified voxel grid sampling}
    \label{algo:downsample}
\end{algorithm}

% about this algorithm, this algorithm was initially added because of the inclusion of per-crack metrics. It might be more complicated if the dnn doesn't produce a 'good enough' results or a very simple one for the opposite. But for sure it is needed either way

\subsection{Deep neural network architecture}
\begin{figure}[t]
    \centering
    \includegraphics[width=.8\linewidth]{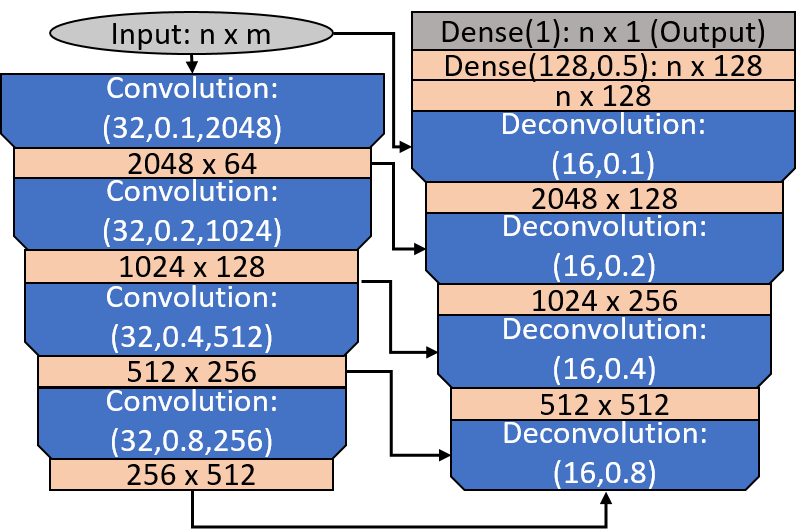}
    \caption{Illustration of PointConv architecture used. \textbf{Convolution: (32,0.1,2048)} is a PointConv layer with neighbourhood $sigma=32$, $radius=0.1$ and $centroids=2048$. \textbf{Deconvolution: (16,0.1)} is a PointDeConv layer with neighbourhood $sigma=32$ and  $radius=0.1$. \textbf{Dense(128,0.5)} is a fully connected layer with output dimension of 128 followed by a dropout layer with a rate of 0.5.}
    \label{fig:pointconv}
\end{figure}
The DNN architecture based on PointConv~\cite{Wu2019} was adapted for the segmentation of crack points because it allows for point-by-point segmentation, and learns directly from input points without the need of converting to another form of representation and computes local region correlation. The architectural layers used is as shown in \cref{fig:pointconv}.

Given a dataset $S = \{V_i, G_i\}$ where $V_i$ is the $i$-th voxel containing $n$ data points, $P_{j=1...n}$, and $G_i$ is the binary ground truth of size $n \times 1$, \{0,1\} corresponding to the class label \{non-crack,crack\} of each data points. The confidence level, $h(P_j|z_j)$ as a crack point is given by the sigmoid function:  
\begin{equation}
    \textstyle h(P_j|z_j) = \frac{1}{1 + exp(-z_j)}
\end{equation} 
\noindent where $z_j$ is the output corresponding to $P_j$ from the final dense layer of the architecture. $h(P_j|z_j)$ has a value [0,1]. To generalise, the notation $H_{\ast}$, \ie $H_i$ for the confidence level of each point in the $i^{th}$ voxel and $H_S$ for dataset $S$, will be used throughout this paper to denote the output from the DNN architecture given the input dataset $S$.

\subsection{Point Clustering}
The output from the DNN only tells us how confident the prediction is on each point in a particular voxel but does not make any inference on whether the points are an instance of a single crack. Grouping points into clusters belonging to the same instance of crack enables the extraction of measurements of the cluster of points \ie crack such as length, width, or volume.

The proposed post-processing follows these steps:
\begin{enumerate*}[label={(\arabic*)}]
    \item Reconstruct surface point clouds from voxels.
    \item For each surface point cloud, take points $p$ with $H_p \geq \Delta_H$ (i.e. points with confidence level greater or equal to $\Delta_H$), where $\Delta_H$ is a pre-defined threshold. Points with $H_p < \Delta_H$ are labelled as non-crack.
    \item The points in $p$ are grouped into clusters, where two points are considered to be from the same cluster if their distance is less than a set threshold $\Delta_r$.
    \item For each cluster on a surface, reject clusters that have fewer points than a set threshold $\Delta_n$ and assign the points as non-cracks. For the remaining clusters, assign a similar value to each point of a cluster, with distinct values across clusters.  
\end{enumerate*} 

\section{Implementation}\label{sec:implementation}
% with the split of dnn using validation still not happy with that, i.e. is it robust and future proof but it actually sounds more reasonable thing to do. Need to discussed this for new stats on the new dataset 
This section describes how PtCrack3D was validated which includes the nature of the dataset, hyperparameters, hardware and the metrics used to evaluate the performance of the proposed crack detection system.

\subsection{Dataset}
To date, there is no publicly available dataset collected for the purpose of evaluating crack detection on an unstructured surface, hence, a new data collection was collected. The Kangaroo Point cliffs (about 18~m high) located in Brisbane, Australia was chosen as the surfaces are unstructured and contain many cracks. A FARO\textsuperscript{\textregistered} Focus\textsuperscript{3D} x330 HDR LIDAR scanner was used to scan the cliffs section-by-section (between 15 to 30~m each). The scanning setting of the LIDAR scanner was set at a constant value of a resolution of 20,480 vertical points per one complete resolution and a scan quality of 6/8. This decision was made due to the scanner's capability (\ie would not allow maxing out both quality and resolution) based on visually inspecting and deciding the balance needed for the point cloud to be able to capture the geometry of cracks. In total there are 21 scans that include the RGB values and the intensity of the reflected beam for each point. Each point was then labelled by hand into two categories: crack and non-crack, hence, a point $j$ contains $P_j\{x_j, y_j, z_j, F_j\{r_j, g_j, b_j, i_j\}\{g_i\}$ where $\{x,y,z\}$ are the position relative to the sensor, $\{r,g,b,i\}$ are the RGB information and intensity respectively, while $g_i$ is the added labelled ground truth.

In total, there are 412 cracks across KP dataset. Two-thirds of the cracks are randomly assigned for training while the rest are for testing. For the training set, surface points that are within 15~cm from the edges of a crack are included as negative. The restriction on the number of surface points included increases the crack points representation to 10\%. The training set is then split randomly into training and validation with a ratio of 2:1. The remaining surface points are included in the testing set. In the testing set, crack points represent 0.08\% of the total points.  

The point clouds are then voxelised, $voxel_{d,n,s}$ using $d=0.5~m$ and $n=2048$. These numbers are chosen based on experimentation where it gives favourable results while balancing the appropriate density of voxels. As for $s$, different values were tested, \mbox{$s=[d, 0.35\times d, 0.30\times d, 0.25\times d]$}. For the testing set, $s$ is set equal to $d$. Voxels with less than $n$ points are discarded from each set. The system was also tested on varying input feature combination. During training, each input point in each voxel is perturbed by $0.001\times r$~m limited to the range $[-0.005,0.005]$ where $r$ is a value randomly sampled from a standard normal distribution.

\begin{figure*}[!t]
    \centering
    \includegraphics[width=0.9\linewidth]{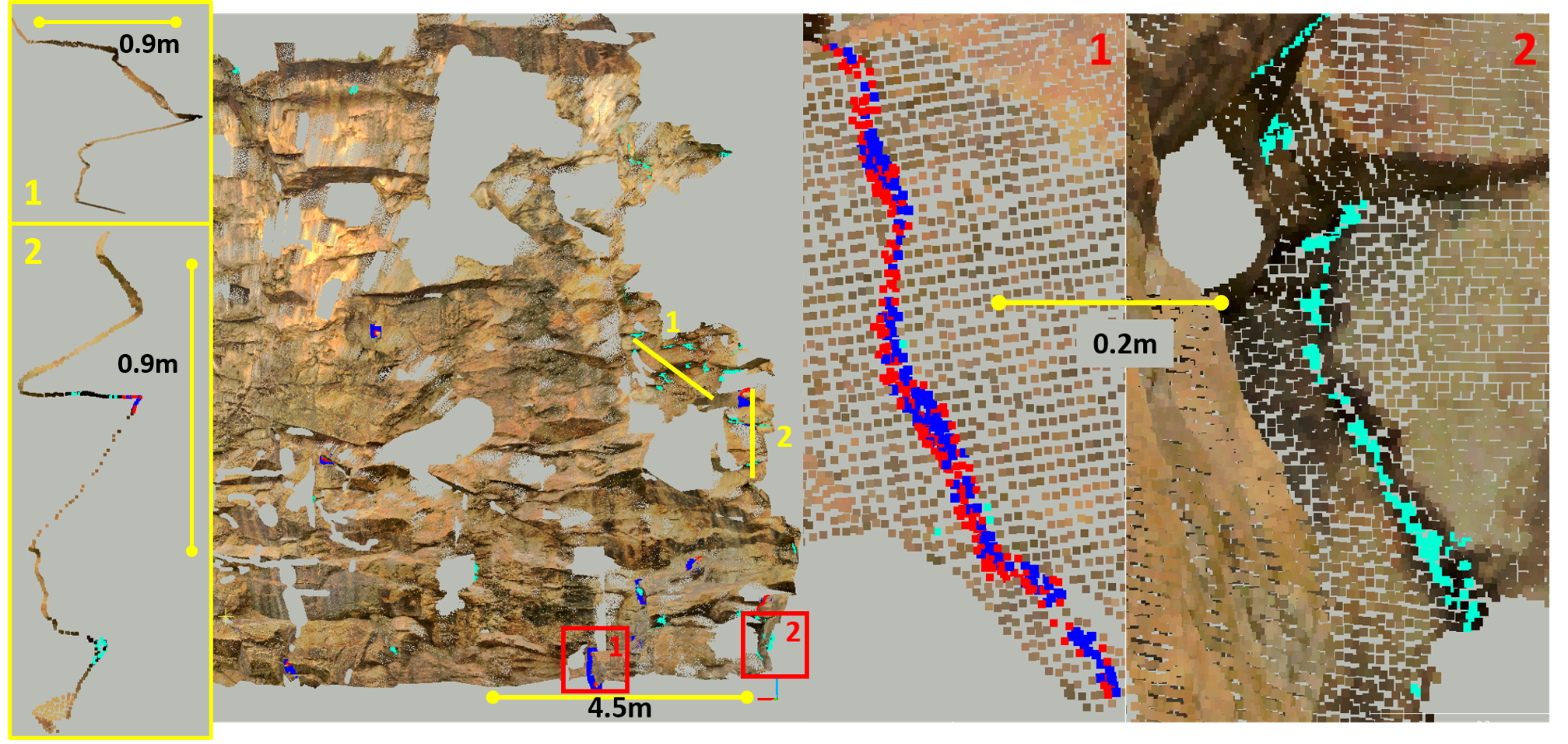}
    \caption{\textbf{Far Left:} Cross section samples of the unstructured surface. \textbf{Left:} Output of PtCrack3D with $\Delta_H=0.59$. Coloured point cloud of a surface with TP (blue), FN (red) and FP (cyan) layered on top. \textbf{Right:} A true crack detection zoomed in. \textbf{Far Right:} A false crack detection zoomed in.}
    \label{fig:outputpc}
\end{figure*}

\subsection{Hyperparamaters}
The hyperparameters used for the PointConv architecture are as shown in Table~\ref{tab:params}. Additionally, the bias of the last layer \ie \textbf{Dense(1)}, of the DNN architecture (fig.~\ref{fig:pointconv}) is initialised to $log(N_{pos}/N_{neg})$ where $N_{pos}$ and $N_{neg}$ is the number of crack points and number of non-crack points in the training dataset respectively. As for the focal loss parameters, all possible combinations of $\gamma$ and $\alpha$ values were tested.
\begin{table}[ht]
    \begin{center}
    \caption{Parameters configuration for the pointConv architecture.}
    \label{tab:params}
    \begin{tabular}{l||l} 
    \textbf{Hyperparameters} & \textbf{Value} \\ \hline \hline
    Optimiser             & Adam ($\beta_{1/2}=(0.9,0.999)$, $\epsilon=10e^{-7}$) \\ \hline
    Epochs                & 101  \\ \hline
    Learning rate, $lr$   & 0.01 \\ \hline
    Learning rate decay   & $lr=0.5\times lr$ after every 10 epochs\\ \hline
    Loss function         & focal loss = ($\gamma, \alpha$)~\cite{lin2017focal}\\ \hline
    Focal Loss parameters & ([2,3,4,5],[10,25,50,75,90]) \\ \hline
    Batch normalisation   & true \\ \hline
    Batch size            & 5 \\ 
    \end{tabular}
    \end{center}
\end{table}

\subsection{Evaluation metrics}
The performance of the output from the DNN segmentation method is evaluated in terms of the per-point precision, recall and specificity score defined respectively as: 
\begin{equation}\label{equ:precall} 
    \begin{gathered}
        \textstyle precision=\frac{TP}{TP+FP} 
        \text{ , }
        \textstyle recall=\frac{TP}{TP+FN} \\
        \textstyle specificity=\frac{TN}{TN+FP}
    \end{gathered}
\end{equation}
% f1 was used in the tables but never defined any where in the text (maybe in-line formula)
% why specificity wasn't used in the per-point classification: (1) the train and validation set are balanced-ish (2)
\noindent where true positives ($TP$) are the number of actual crack points classified as crack, false negatives ($FN$) are the number of actual crack points classified as non-cracks and false positives ($FP$) are the number of surface points classified as cracks. These metrics give the performance of the detector at the per-point level, however, it does not capture the performance of the detector at per-crack level (crack instance detection). Hence, we define the rate of crack instances detected, $cr_{det}$ as:
\begin{equation}
    cr_{det}=\frac{1}{N_{cr}}\sum_{n=1}^{N_{cr}}
    \begin{cases}
        1, & \text{if any } cr_{pred} \cap cr_{real}\\
        0, & \text{otherwise}
    \end{cases}
    \label{equ:detection}
\end{equation}
\noindent where $N_{cr}$ is the total number of cracks in the dataset, $cr_{pred}$ is an instance of detected crack and $cr_{real}$ is an instance of a real crack. In Eq.~\ref{equ:detection}, the condition $cr_{pred} \cap cr_{real}$ (\ie a detected crack instance intersects a real crack instance) is only true if the \{number of intersected points\} $\geq$ \{$\alpha*$number of points in $cr_{pred}$\}, where $\alpha$ is a number between (0,1]. It is also true if one or more $cr_{pred}$ intersects one $cr_{real}$, hence this metric is paired with crack continuity:
\begin{equation}
    cr_{con}=\textstyle\frac{1}{N_{cr}}\sum_{n=1}^{N_{cr}}\frac{1}{N_{n,cr_{pred} \cap cr_{real}}}
    \label{equ:continuity}
\end{equation}
\noindent where $N_{cr}$ is the total number of real cracks in the dataset and $N_{n,cr_{pred} \cap cr_{real}}$ is the number of predicted crack instance that satisfies the condition $cr_{pred} \cap cr_{real}$ for the $n^{th}$ crack. This metric measures the fragmentation of the detection on a crack. Finally crack precision, $cr_{pre}$ is defined as:
\begin{equation}
    cr_{pre}=\textstyle\frac{N_{cr_{pred} \cap cr_{real}}}{N_{cr_{pred}}}
    \label{equ:crackprecision}
\end{equation} 
where $N_{cr_{pred} \cap cr_{real}}$ is the number of predicted crack instance that satisfies the condition $cr_{pred} \cap cr_{real}$ and $N_{cr_{pred}}$ is the total number of predicted crack instances in the dataset. This metric gives the ratio of crack clusters detected to the total clusters detected. Each metrics generates a value between [0,1] where a value closer to one indicates better performance. Note that these metrics can only be evaluated after the post-processing step as the output from this step are clusters of points where each cluster is considered as an instance of detected cracks.

\subsection{Training Execution}

PtCrack3D was trained and tested on an Intel i7-8700K, 64GB RAM with NVIDIA GeForce GTX 1080 Ti graphics card. The training set used was augmented by translating the point cloud 10 times along randomly picked axes by a value randomly generated from within [0, 0.50]~m. 
%This ensures having enough data for generalisation during training. 
The results shown in this section are based on the validation set. The goal of the experimentation was to determine the best DNN configuration and parameters for the crack detection system.

The first test was to discover the best $\gamma$ and $\alpha$ values for the focal loss function. For this test, the position $(x,y,z)$ of the point was used as the only feature input. While a number of combinations were used (as in \cref{tab:params}), only tests showing promising results were included in \cref{tab:re:focal}. 
%To evaluate which parameter configurations were the most desirable, 
The configuration with the highest F1 score was chosen. 
To further justify the use of focal loss and the chosen parameters, the focal loss result was compared to the result using binary cross-entropy as shown in \cref{tab:re:focal} -- last row. The F1 score and recall value of the focal loss was higher than of the binary cross-entropy, however, binary cross-entropy performed better in terms of precision which is common when there is a class imbalance towards negative samples. By applying focal loss, during training, less weight is given to easy sample points \ie non-crack and vice versa. For the application of crack detection where failure to detect cracks could lead to disastrous consequences, a higher recall value is more desirable, whilst compromising the precision.  

\begin{table}[!ht]
    \caption{Point-wise performance with varying focal loss parameters.}
    \label{tab:re:focal}
    \begin{center}
        \begin{tabularx}{.85\columnwidth}{Y|Y||c|c|c} 
            \textbf{$\gamma$} & \textbf{$\alpha$} & \textbf{Precision} & \textbf{Recall} & \textbf{F1 score }\\ \hline \hline
            3 & 0.75    & 0.32 & 0.23 & 0.27 \\ \hline
            3 & 0.90    & 0.25 & 0.68 & 0.37 \\ \hline
            \rowcolor{green} 4 & 0.75 & 0.39 & 0.41 & 0.40 \\ \hline
            4 & 0.90    & 0.24 & 0.66 & 0.35 \\ \hline
            5 & 0.75    & 0.32 & 0.26 & 0.29 \\ \hline
            5 & 0.90    & 0.22 & 0.64 & 0.33 \\ \hline
            \multicolumn{2}{c||}{Binary cross entropy} & 0.64 & 0.21 & 0.31 \\
        \end{tabularx} 
    \end{center}
\end{table}

We evaluated the effect of using different combination of point features, see \cref{tab:re:feature}. The results show that adding colour information ($rgb$) on top of position-only ($xyz$) increases the performance in both precision and recall by about 10\% and 20\% respectively. This is because the structure of cracks often makes them appear darker than their surrounding surface. Adding the LIDAR intensity $i$ to $xyz$ increases performance by almost 5\%, however, adding $i$ on top of $rgb$ does not seem to alter the performance as it is likely that $rgb$ already captures the information useful in discriminating cracks. 
%Adding $i$ on top of $rgb$ would just add redundant information that already exists in the $rgb$ values. 

\begin{table}[!ht]
    \caption{Point-wise detection results with varying feature input.}
    \label{tab:re:feature}
    \begin{center}
    \begin{tabular}{c||c|c|c} 
    \textbf{Feature input, s (m)} & \textbf{Precision} & \textbf{Recall} & \textbf{F1 score }\\ \hline \hline
    $x,y,z$          & 0.39 & 0.41 & 0.40 \\ \hline
    $x,y,z,i$        & 0.43 & 0.45 & 0.44 \\ \hline
    \rowcolor{green} $x,y,z,r,g,b$    & 0.50 & 0.60 & 0.55 \\ \hline
    $x,y,z,r,g,b,i$  & 0.50 & 0.60 & 0.55 \\ 
    \end{tabular}
    \end{center}
\end{table}

%The final test was done by varying the stride length during voxelisation. 
Finally we tested the impact of varying stride lengths 
%ranging from no overlapping voxels, $s=d$ to 75\% overlapping voxels, $s=0.125$. 
%For this test, 
using the original training set \ie without the translation augmentation, see \cref{tab:re:voxel}. Increasing the overlapping increases the performance in all aspects. This is due to the network having more chance of learning a certain feature multiple times in different positions in a voxel. However, having lower stride length increases the number of input voxels, hence increasing the time required to complete the training of 101 epochs. As in \cref{tab:re:voxel} -- 3 last rows, an increase of 5\% overlap caused the training duration to almost double. 

\begin{table}[!ht]
    \caption{Point-wise detection performance with varying voxel stride.}
    \label{tab:re:voxel}
    \begin{center}
    \begin{tabularx}{.95\columnwidth}{Y||c|c|c|Y} 
    \textbf{voxel stride, s (m) [\% overlap]} & \textbf{Precision} & \textbf{Recall} & \textbf{F1 score } & \textbf{Duration (d,h,m)}\\ \hline \hline
    0.500 [0\%] & 0.45 & 0.47 & 0.46 & 0,02,29 \\ \hline
    0.250 [50\%] & 0.50 & 0.50 & 0.50 & 0,15,37 \\ \hline
    0.175 [65\%] & 0.54 & 0.60 & 0.57 & 1,10,38 \\ \hline
    0.150 [70\%] & 0.58 & 0.63 & 0.60 & 2,07,45 \\ \hline
    \rowcolor{green} 0.125 [75\%] & 0.60 & 0.67 & 0.63 & 4,02,31\\
    \end{tabularx}
    \end{center}
\end{table}
\section{Experimental Results}\label{sec:results}

The detection performance of PtCrack3D was evaluated using the model trained with voxel stride of 0.125~m, 
%The results from the point-wise and crack-wise detection are discussed as follows:
and compared to SOTA defects detection, MPFH~\cite{madrigal2017} as mentioned in~\cref{sec:litreview}. For a fair comparison, MPFH was fine-tuned to our dataset. For our MPFH implementation, a neural network was used after the feature extraction step. The $rgb$ values of each point were also included as a feature on top of the extracted ones, adding 3 additional features for the neural net. This addition of RGB values increased the overall performance of MPFH with an absolute increase of around 7\% for precision and recall during training.

\subsection{Pointwise Classification}
\begin{figure}[htp]
    \centering
    \includegraphics[width=0.9\linewidth]{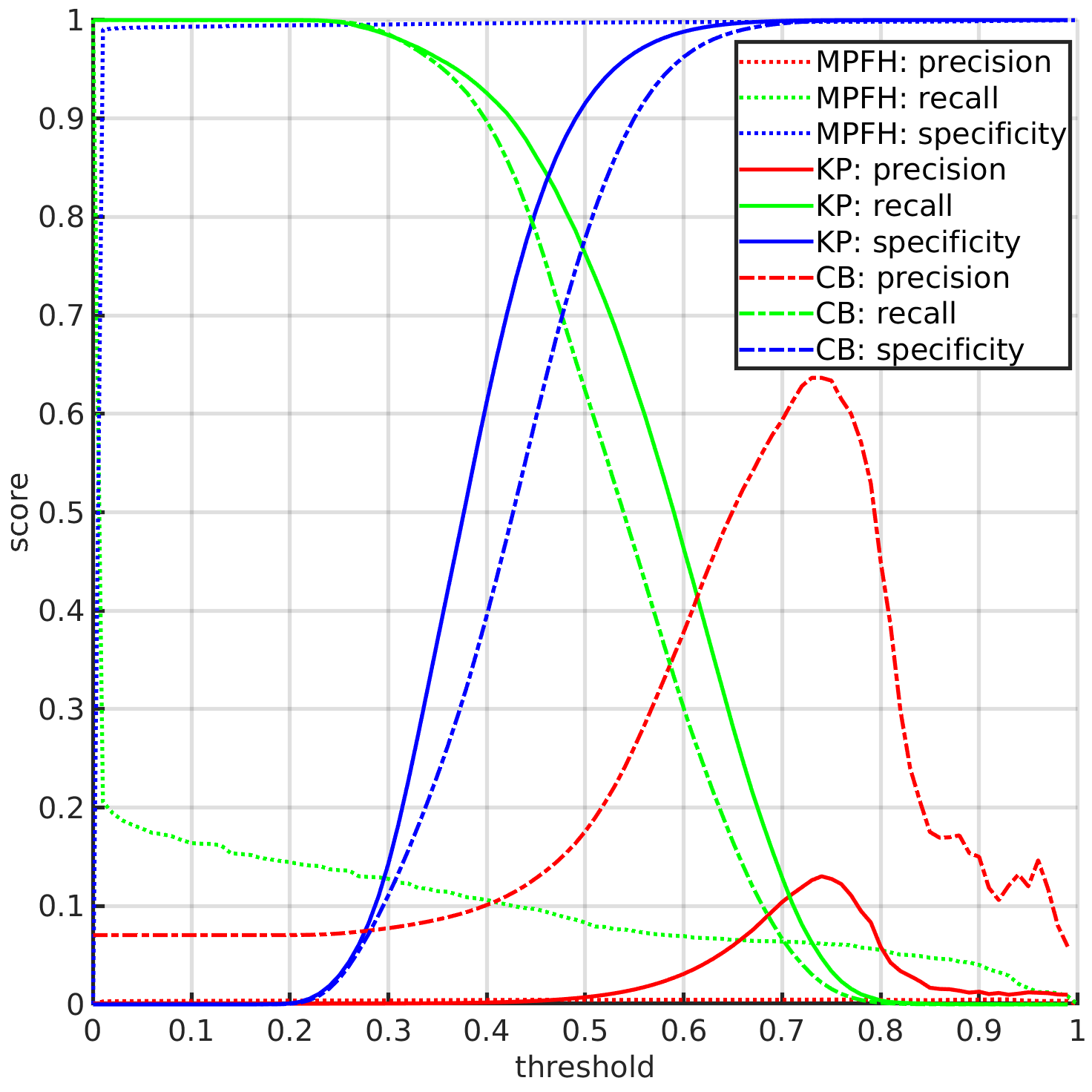} %{imgs/pre-rec-f1-all.eps}
    \caption{Performance curves of PtCrack3D on Kangaroo Point (KP) and Coolum Beach (CB) dataset and MPFH applied on KP.}
    \label{fig:prerecspe}
\end{figure}

\cref{fig:prerecspe} shows PtCrack3D per-point classification performance (in solid lines). At threshold, $\Delta_H = 0.5$, PtCrack3D obtained a recall score of 76\%, specificity of 92\% and a precision of 0.7\%. The low precision score is due to the large class imbalance where small percentage of surface points falsely detected as crack points will exponentially decrease the precision score.

In comparison, MPFH, shown in dotted-lines in~\cref{fig:prerecspe} at $\Delta_H = 0.5$, achieved a recall score of 8\%, which is 68\% lower than PtCrack3D in absolute. The precision is consistent at 0.5\% across all threshold values while ours scored 13\% maximum when the threshold is $\Delta_H=0.74$.  

\subsection{Crack Detection}
% nope, on the model trained with point cloud data with voxel stride of XXXX
%and the configuration of the green-highlighted row on \cref{tab:re:voxel} was used. 
This section focuses on the detection of crack instances, after the post-processing step of PtCrack3D. \cref{tab:re:postpro} shows the results for different confidence thresholds and post-processing configurations. 
For our method, we found that $\Delta_H=0.59$ provides the most relevant balance between precision and recall, given the application. In this configuration the method detected 97\% of the cracks in the test dataset, even though only half of the crack points were classified as belonging to a crack. 
%increasing the $\Delta_H$ up to 0.59 still managed to detect 97\% of cracks even when the system only detects half of crack points as crack. 
For MPFH, the results are shown for $\Delta_H=0.005$ and $[\Delta_r=0.04,\Delta_n=15]$ where the performance was best (both precision and recall are at their maximum). MPFH was only able to detect 47\% of the cracks instances. 
%it gives the best performance overall. 
PtCrack3D clearly outperforms MPFH, even with the addition of colour, across all metrics. 
Increasing the threshold $\Delta_H$ to 0.65 for our method, leads to a similar recall to that obtained with MPFH, but with a much higher precision and still a significantly higher crack detection rate of 78\%. 

\begin{figure*}[htp]
    \centering
    \includegraphics[width=0.8\linewidth]{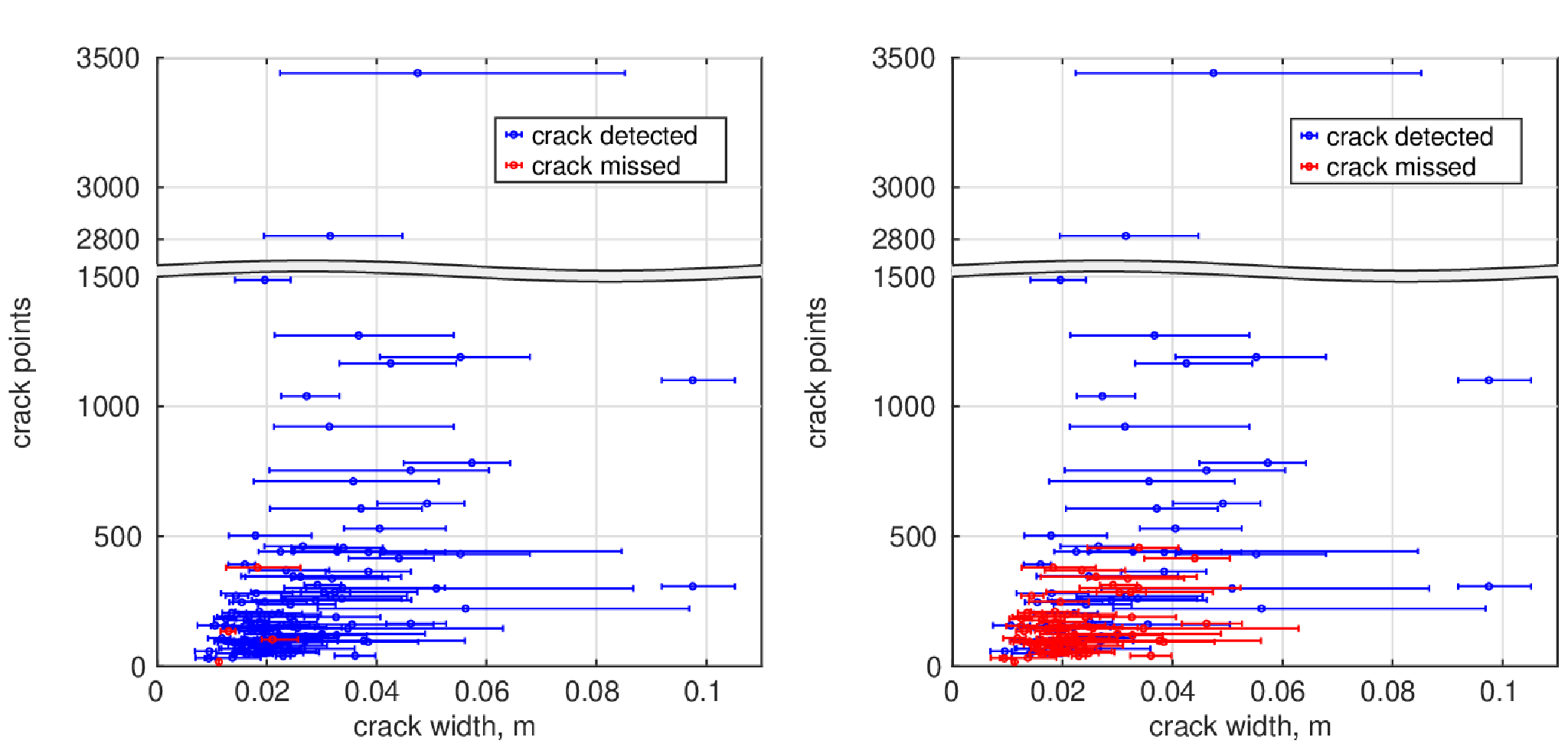} %hist-width.jpg
    \caption{Crack detection output in relation to crack width and number of crack points for each individual crack in the test set (dot is the mean width while the ends are the width extremities). \textbf{Left:} PtCrack3D. \textbf{Right:} MPFH.}
    \label{fig:width}
\end{figure*}
%For comparison to our system, a $\Delta_H$ = 0.65 was used due to the close recall and specificity scores. Even with similar per-point scores, 
%our system performed better overall and managed to detect 78\% of crack compared to only 47\% of MPFH.

 % Looking at precision, recall and specificity score, the precision value increased up to 10\% across each configuration without any significant compromise to the recall values. This was because the post-processing step corrects the false positives as these points tended to be in small clusters on the surface. Focusing on $\Delta_H = 0.512$ (\ie the threshold of where the precision and recall value intersects as in~\cref{fig:prerecspe}), for the crack specific metrics, the proposed algorithm was able to almost correctly identify all cracks that existed on the surfaces. Most of these real cracks are detected as a single crack given by the high $C_{con}$ score. Out of all the detected clusters, 75\% of the clusters are true positive.

\begin{table}[b]
    \centering
    \caption{Performance of crack detection methods. $\Delta_H$ = Confidence threshold, [$\Delta_r$, $\Delta_n$] = [distance threshold, point threshold].}
    \begin{tabularx}{\columnwidth}{c||c|c|c|c}
    \multicolumn{1}{c||}{} & \multicolumn{3}{c|}{PtCrack3D}  & \multicolumn{1}{Y}{MPFH} \\ \hline
    $\Delta_H$  & 0.50 & \textbf{0.59} & 0.65 & 0.005 \\ \hline
    [$\Delta_r$, $\Delta_n$]  & [0.02,20] & \textbf{[0.04,20]} & [0.04,15] & [0.04,15] \\ \hhline{=====}
    precision    & 0.013   & \textbf{0.05}  & 0.15 & 0.023 \\ \hline
    recall       & 0.73    & \textbf{0.49}  & 0.26 & 0.29 \\ \hline
    specificity  & 0.95    & \textbf{0.99}  & 0.99 & 0.99 \\ \hline
    $cr_{det}$   & 0.99    & \textbf{0.97}  & 0.78 & 0.47 \\ \hline
    $cr_{con}$   & 0.81    & \textbf{0.92}  & 0.84 & 0.83 \\ \hline
    $cr_{pre}$   & 0.0075  & \textbf{0.027} & 0.084 & 0.02 \\
    \end{tabularx}
    \label{tab:re:postpro}
\end{table}
% the different choices of \Delta_r,n have to be clearly stated, is it justified by saying that these deltas are not kept constant to give a fair comparison between different \Delta_H by maximising the performance.
% for further analysis (prob in thesis) have results of varying \Delta_r,n on a specific \Detla_H
% also explanation of the changes in per-point precision, recall, and specificity after post-processing 
%In general, decreasing $\Delta_H$ increases the crack detection rate $cr_{det}$ and continuity $cr_{con}$ scores. 
%Further decreasing the $\Delta_H$ will produce a crack cluster that overlaps most of the surface points (i.e. detecting the whole surface as a crack). 
%Increasing $\Delta_H$ decreases $cr_{det}$ and $cr_{con}$ but reduces the number of false positive clusters given by the increasing $cr_{pre}$ value.   
Further evaluation of our method (with $\Delta_H=0.59$) was performed by analysing the detection outcome against the width and the number of points of each crack instance, see~\cref{fig:width}. The width of a crack was measured by taking the shortest distances between the points on the opposite edges of the crack at a constant interval across its length. This was done manually from the labelled point cloud. 
\cref{fig:width} shows that the proposed method successfully detected 100\% of the cracks with a maximum width of \SI{0.03}{m} or above. In contrast, MPFH detected 91\% of the 11 cracks with a maximum width above \SI{0.06}{m} and only 61\% of the 57 cracks with a maximum width above \SI{0.03}{m}.

PtCrack3D was able to detect 100\% of crack with more than 500 points, which is similar to MPFH. As for cracks with 500 points and below, PtCrack3D managed to detect 97\% of them while MPFH only detected 40\%. Having more points for a small crack allows for more relevant information to be extracted compared to small cracks with fewer points. 

At $\Delta_H = 0.65$ and $0.005$ for PtCrack3D and MPFH, where the per-point recall scores are almost similar, PtCrack3D was able to detect more cracks compared to MPFH. This is because our method is more consistent in the ability to detect crack points across each crack (i.e. being able to detect about the same percentage of points across all cracks) while MPFH detects more points on easier cracks (i.e. cracks with larger width or points) and failed to detect any points on harder cases. 
% this paragraph is referring to when proposed method and MPFH \Delta_H = 0.65, 0.005 respectively which the per-point recall are almost the same. Have to specify this as it is not clear in this paragraph i.e. to which configuration was it compared to  

Qualitative analysis of the results shows that our system was able to detect cracks with sufficient points for characterisation. As shown in~\cref{fig:outputpc}, where the coloured points \ie blue, red and cyan are TP, FN and FP crack points classified by the method. The FN are scattered in groups of small numbers where the blue clusters are sufficient to extract crack characteristics such as length. However, PtCrack3D also has the tendency to detect sharp-concaved edges as a crack as can be seen in numbers on the right-hand side of the full surface in the left-most image of~\cref{fig:outputpc} with a close-up view of a sample on the left-most image.   

\subsection{Cross-validation}

A second dataset was collected to test the trained model on a different scene. Scans of unstructured rock surfaces of Coolum Beach (see \cref{fig:scanning}), located \SI{120}{km} north of Kangaroo Point, were captured using the same LIDAR scanner and configuration. The rock formations are between \SI{1.5}-\SI{3}{m} in height and \SI{3}-\SI{5}{m} in width which allows for high density point cloud across the surfaces. In total, 5 scans were captured covering a total surface area of approximately more than \SI{100}{m^2} with 35 cracks representing about 7\% of the total points with width ranging from \SI{0.5}{cm} to \SI{35}{cm}. 

\subsubsection{Implementation}

PtCrack3D was applied on the entire Coolum dataset without any re-training, with the same implementation and parameters as in \cref{sec:implementation} except for the post-processing step. The $[\Delta_r, \Delta_n]$ parameters of the point clustering were set to $[0.05,300]$ to account for the larger dimensions of the cracks and higher point density due to the sensor being closer to the scanned surface. 

\subsubsection{Results and Discussions}
PtCrack3D managed to detect all of the cracks in the dataset in terms of crack-wise performance. Note that the smallest crack is about \SI{0.5}{cm} and \SI{0.7}{cm} in width at the smallest and largest opening, respectively. The high resolution of the smallest crack (\ie length and points about \SI{12}{cm} and 500 points, respectively) enables more geometrical information available for detection. For the detected cracks, PtCrack3D achieved a performance of $cr_{con} = 0.94$ and $cr_{pre} = 0.71$ which is on par with the Kangaroo Point dataset. 

\begin{figure}[htp]
    \centering
    \includegraphics[width=.9\linewidth]{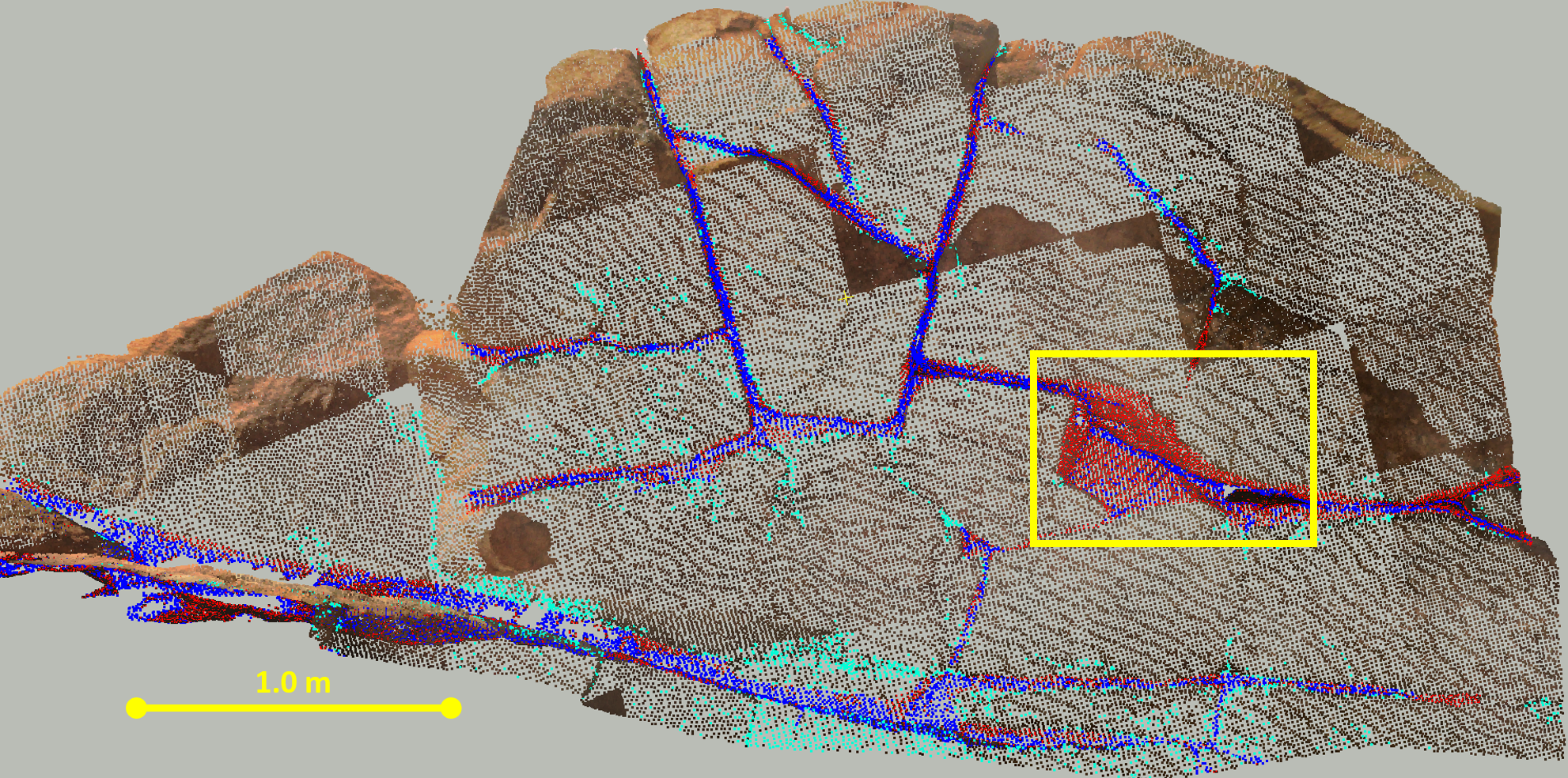} 
    \caption{Coolum dataset sample surface point cloud output of PtCrack3D. Coloured point cloud of a surface with TP (blue), FN (red) and FP (cyan).}
    \label{fig:cb_fin}
\end{figure}

\cref{fig:cb_fin} shows a visualisation of the point cloud output on a surface sample from the Coolum dataset which represents the detection behaviour in general. The FPs (in cyan) are similar to the behaviour on Kangaroo Point dataset \ie mostly around the edges of cracks and in corners with high changes of surface normals. Achieving to detect all crack instances, PtCrack3D underestimates the width of a crack with a large width as highlighted in the yellow box of \cref{fig:cb_fin}. This is because the opening width is more than \SI{35}{cm} at the widest point, whereas the largest opening on the dataset used to train PtCrack3D is only about \SI{10}{cm}.

\cref{fig:prerecspe} shows the point-wise performance curves (in dash-dotted lines) for the trained PtCrack3D method implementation on the Coolum dataset. A small absolute drop in recall and specificity (\ie at threshold of 0.5) of about 14\% can be observed compared to the Kangaroo Point test performance. The precision increased by about 20\%. However, this increase in performance reflects the higher crack to non-crack point ratio of the Coolum dataset. Normalising the Coolum precision score by equalising its FP rate to the Kangaroo Point performance, shows an absolute drop of about 18\% compared to the expected precision scores given by the differences in crack points ratio. The drop in performance across all measured point-wise metrics is expected as the model are trained using a different dataset. As mentioned earlier, it is more important to be able to detect cracks as an object, whereas the point-wise detection enables fine details measurements to be extracted from detected cracks.

\section{Conclusions}\label{sec:conclusion}

In this paper, we proposed PointCrack3D, a point-cloud-based crack detection method capable of automatically detecting crack instances from 3D LIDAR data in highly unstructured surfaces such as rock cliffs or natural/mine tunnels. This is the first crack detection method utilising a point-based deep learning architecture for detection on unstructured surfaces. 
PtCrack3D includes a method which down-samples a point cloud while maintaining the point density of cracks and a post-processing step to segment cracks via clustering of 3D points corresponding to the detected crack instances. 
%As shown in \cref{fig:outputpc} the post-processing step clusters points into instances of crack whilst reducing the number of false-positive points. 
%PtCrack3D is particularly relevant in applications where detecting instances of individual cracks is important, such as geo-technical analysis. 

PtCrack3D was first experimentally validated on a dataset of LIDAR point clouds capturing a large natural rock cliff. The method was shown to detect 97\% of the crack instances in the test set overall, and 100\% of cracks with a maximum width of more than \SI{3}{cm}, where the SOTA MPFH defects detection method could only detect 47\% (and 61\%, resp.) of those cracks.
PtCrack3D was then applied `out of the box' to a second dataset acquired in a distinct location
%where the smallest crack measures about \SI{0.6}{cm} in width, 
and managed to detect all cracks, indicating its detection capability on an out-of-sample case.
The method was evaluated in different parameters configuration that could be fine-tuned to fit detection requirements. As an example, setting a high confidence threshold value may be suitable for applications where small cracks are less important to be identified whilst focusing on larger cracks and keeping false alarms low. 
It could also be concluded that the detection rate depends both on the width of a crack and the number of points that makes up the crack. 
Automatically detecting cracks in highly unstructured surfaces using PtCrack3D is a critical step towards the deployment of robots for autonomous crack inspection in an unstructured environment.

%%%%%%%%%%%%%%%%%%%%%%%%%%%%%%%%%%%%%%%%%%%%%%%%%%%%%%%%%%%%%%%%%%%%%%%%%%%%%%%%
% \addtolength{\textheight}{-12cm}   % This command serves to balance the column lengths
%                                   % on the last page of the document manually. It shortens
%                                   % the textheight of the last page by a suitable amount.
%                                   % This command does not take effect until the next page
%                                   % so it should come on the page before the last. Make
%                                   % sure that you do not shorten the textheight too much.

%%%%%%%%%%%%%%%%%%%%%%%%%%%%%%%%%%%%%%%%%%%%%%%%%%%%%%%%%%%%%%%%%%%%%%%%%%%%%%%%
% \input{8_appendix.tex}
% \input{9_acknow.tex}

%%%%%%%%%%%%%%%%%%%%%%%%%%%%%%%%%%%%%%%%%%%%%%%%%%%%%%%%%%%%%%%%%%%%%%%%%%%%%%%%

\bibliographystyle{IEEEtran.bst}
\bibliography{IEEEabrv.bib,references-fin.bib}

\begin{thebibliography}{10}
\providecommand{\url}[1]{#1}
\csname url@rmstyle\endcsname
\providecommand{\newblock}{\relax}
\providecommand{\bibinfo}[2]{#2}
\providecommand\BIBentrySTDinterwordspacing{\spaceskip=0pt\relax}
\providecommand\BIBentryALTinterwordstretchfactor{4}
\providecommand\BIBentryALTinterwordspacing{\spaceskip=\fontdimen2\font plus
\BIBentryALTinterwordstretchfactor\fontdimen3\font minus
  \fontdimen4\font\relax}
\providecommand\BIBforeignlanguage[2]{{%
\expandafter\ifx\csname l@#1\endcsname\relax
\typeout{** WARNING: IEEEtran.bst: No hyphenation pattern has been}%
\typeout{** loaded for the language `#1'. Using the pattern for}%
\typeout{** the default language instead.}%
\else
\language=\csname l@#1\endcsname
\fi
#2}}

\bibitem{molinda2000}
G.~M. Molinda, C.~Mark, and D.~R. Dolinar, ``{Assessing coal mine roof
  stability through roof fall analysis},'' \emph{Proc. of the new technology
  for coal mine roof support}, vol. 9453, no. 151, pp. 53--72, 2000.

\bibitem{szwedzicki2003}
T.~Szwedzicki, ``{Rock mass behaviour prior to failure},'' \emph{Int. J. of
  Rock Mechanics and Mining Sciences}, vol.~40, no.~4, pp. 573--584, 2003.

\bibitem{msha2020}
\BIBentryALTinterwordspacing
{MSHA}, ``{Mine Injury and Worktime Reports},'' 2020. [Online]. Available:
  \url{https://arlweb.msha.gov/ACCINJ/accinj.htm}
\BIBentrySTDinterwordspacing

\bibitem{Chen2020}
F.-C. Chen and M.~R. Jahanshahi, ``{ARF-Crack: rotation invariant deep fully
  convolutional network for pixel-level crack detection},'' \emph{Machine
  Vision and Applications}, vol.~31, no.~6, p.~47, 2020.

\bibitem{azhari2019}
F.~Azhari, C.~Sennersten, and T.~Peynot, ``{Evaluation of Vision-based Surface
  Crack Detection Methods for Underground Mine Tunnel Images},'' in
  \emph{Australas. Conf. on Robotics and Automation}, 2019.

\bibitem{Dung2019}
C.~V. Dung and L.~D. Anh, ``{Autonomous concrete crack detection using deep
  fully convolutional neural network},'' \emph{Automation in Construction},
  vol.~99, pp. 52--58, 2018.

\bibitem{zhu2019}
Q.~Zhu, M.~D. Phung, and Q.~Ha, ``{Crack Detection Using Enhanced Hierarchical
  Convolutional Neural Networks},'' in \emph{Australas. Conf. on Robotics and
  Automation}, 2019.

\bibitem{zou2019}
Q.~Zou, Z.~Zhang, Q.~Li, X.~Qi, Q.~Wang, and S.~Wang, ``{DeepCrack: Learning
  Hierarchical Convolutional Features for Crack Detection},'' \emph{{IEEE}
  Trans. Image Processing}, vol.~28, no.~3, Mar. 2019.

\bibitem{Montero2017}
R.~Montero, E.~Menendez, J.~G. Victores, and C.~Balaguer, ``{Intelligent
  robotic system for autonomous crack detection and characterization in
  concrete tunnels},'' in \emph{ICARSC}, Apr. 2017, pp. 316--321.

\bibitem{Zhong2020}
M.~Zhong, L.~Sui, Z.~Wang, and D.~Hu, ``{Pavement Crack Detection from Mobile
  Laser Scanning Point Clouds Using a Time Grid},'' \emph{Sensors}, vol.~20,
  no.~15, July 2020.

\bibitem{Zhang2019-crack}
A.~Zhang, K.~C.~P. Wang, Y.~Fei, Y.~Liu, C.~Chen, G.~Yang, J.~Q. Li, E.~Yang,
  and S.~Qiu, ``{Automated Pixel-Level Pavement Crack Detection on 3D Asphalt
  Surfaces with a Recurrent Neural Network},'' \emph{Computer-Aided Civil and
  Infrastruct. Eng.}, vol.~34, no.~3, pp. 213--229, Mar. 2019.

\bibitem{Xu2019}
X.~Xu and H.~Yang, ``{Intelligent crack extraction and analysis for tunnel
  structures with terrestrial laser scanning measurement},'' \emph{Advances in
  Mech. Eng.}, vol.~11, no.~9, Sept. 2019.

\bibitem{sarker2017}
M.~M. Sarker, T.~A. Ali, A.~Abdelfatah, S.~Yehia, and A.~Elaksher, ``{A cost
  effective method for crack detection and measurement on concrete surface},''
  \emph{ISPRS}, vol. XLII-2/W8, pp. 237--241, 2017.

\bibitem{Cho2018}
S.~Cho, S.~Park, G.~Cha, and T.~Oh, ``{Development of Image Processing for
  Crack Detection on Concrete Structures through Terrestrial Laser Scanning
  Associated with the Octree Structure},'' \emph{Appl. Sci.}, vol.~8, no.~12,
  Nov. 2018.

\bibitem{cabaleiro2017}
M.~Cabaleiro, R.~Lindenbergh, W.~Gard, P.~Arias, and J.~van~de Kuilen,
  ``{Algorithm for automatic detection and analysis of cracks in timber beams
  from LiDAR data},'' \emph{Constr. Build. Mater.}, vol. 130, Jan. 2017.

\bibitem{jovancevic2017}
I.~Jovan{\v{c}}evi{\'{c}}, H.-H. Pham, J.-J. Orteu, R.~Gilblas, J.~Harvent,
  X.~Maurice, and L.~Br{\`{e}}thes, ``{3D Point Cloud Analysis for Detection
  and Characterization of Defects on Airplane Exterior Surface},'' \emph{J. of
  Nondestruct. Eval.}, vol.~36, no.~74, 2017.

\bibitem{Allen2018}
Z.~Allen, W.~K.~C. P., F.~Yue, L.~Yang, T.~Siyu, C.~Cheng, L.~J. Q., and
  L.~Baoxian, ``{Deep Learning–Based Fully Automated Pavement Crack Detection
  on 3D Asphalt Surfaces with an Improved CrackNet},'' \emph{J. of Computing in
  Civil Eng.}, vol.~32, no.~5, Sept. 2018.

\bibitem{madrigal2017}
C.~Madrigal, J.~Branch, A.~Restrepo, and D.~Mery, ``{A Method for Automatic
  Surface Inspection Using a Model-Based 3D Descriptor},'' \emph{Sensors},
  vol.~17, no.~10, Oct. 2017.

\bibitem{Charles2017}
R.~Q. Charles, H.~Su, M.~Kaichun, and L.~J. Guibas, ``{PointNet: Deep Learning
  on Point Sets for 3D Classification and Segmentation},'' in \emph{CVPR},
  2017, pp. 77--85.

\bibitem{charles2017-pp}
C.~R. Qi, L.~Yi, H.~Su, and L.~J. Guibas, ``{PointNet++: Deep Hierarchical
  Feature Learning on Point Sets in a Metric Space},'' 2017.

\bibitem{Hua2018}
B.-S. Hua, M.-K. Tran, and S.-K. Yeung, ``{Pointwise Convolutional Neural
  Networks},'' in \emph{CVPR}, June 2018.

\bibitem{Wu2019}
W.~Wu, Z.~Qi, and L.~Fuxin, ``{PointConv: Deep Convolutional Networks on 3D
  Point Clouds},'' in \emph{CVPR}, 2019, pp. 9613--9622.

\bibitem{li2018-pointcnn}
Y.~Li, R.~Bu, M.~Sun, and B.~Chen, ``{PointCNN: : Convolution On X-Transformed
  Points},'' \emph{arXiv:1801.07791}, 2018.

\bibitem{randlanet}
Q.~Hu, B.~Yang, L.~Xie, S.~Rosa, Y.~Guo, Z.~Wang, N.~Trigoni, and A.~Markham,
  ``{RandLA-Net: Efficient Semantic Segmentation of Large-Scale Point
  Clouds},'' in \emph{CVPR}, 2020.

\bibitem{cylinandasym}
X.~Zhu, H.~Zhou, T.~Wang, F.~Hong, Y.~Ma, W.~Li, H.~Li, and D.~Lin,
  ``{Cylindrical and Asymmetrical 3D Convolution Networks for LiDAR
  Segmentation},'' in \emph{CVPR}, 2021.

\bibitem{fgnet}
K.~Liu, Z.~Gao, F.~Lin, and B.~M. Chen, ``{FG-Net: Fast Large-Scale LiDAR Point
  CloudsUnderstanding Network Leveraging CorrelatedFeature Mining and
  Geometric-Aware Modelling},'' \emph{arXiv:2012.09439}, 2020.

\bibitem{Soilan2019}
M.~Soil{\'{a}}n, R.~Lindenbergh, B.~Riveiro, and
  A.~S{\'{a}}nchez-Rodr{\'{i}}guez, ``{PointNet for the automatic
  classification of aerial point clouds},'' \emph{ISPRS}, vol. IV-2/W5, pp.
  445--452, May 2019.

\bibitem{Li2020}
J.~Li, L.~Fu, P.~Wang, and C.~Sun, ``{Indoor point cloud recognition with deep
  convolutional networks},'' in \emph{Int. Conf. on Opt. Instrum. and
  Technol.}\hskip 1em plus 0.5em minus 0.4em\relax SPIE, 2020, pp. 65--74.

\bibitem{Ma2019}
L.~Ma, Y.~Li, J.~Li, W.~Tan, Y.~Yu, and M.~A. Chapman, ``{Multi-Scale
  Point-Wise Convolutional Neural Networks for 3D Object Segmentation From
  LiDAR Point Clouds in Large-Scale Environments},'' \emph{{IEEE} Trans.
  Intell. Transport. Syst.}, 2020.

\bibitem{lin2017focal}
T.-Y. Lin, P.~Goyal, R.~Girshick, K.~He, and P.~Doll{\'{a}}r, ``{Focal Loss for
  Dense Object Detection},'' in \emph{ICCV}, Oct. 2017, pp. 2999--3007.

\end{thebibliography}

\end{document}